\documentclass[10pt, conference, compsocconf]{IEEEtran}
\usepackage{microtype}
\usepackage{graphicx}
\usepackage{subfigure}
\usepackage{amsmath,amsthm,amssymb}
\usepackage{booktabs} 
\usepackage{physics}

\ifCLASSINFOpdf
\else
\fi
\usepackage{algorithmic}
\usepackage{url}

\usepackage{hyperref}

\usepackage{eso-pic}

\begin{document}
\AddToShipoutPictureBG*{%
  \AtPageUpperLeft{%
    \hspace{.73\paperwidth}%
    \raisebox{-\baselineskip}{%
      \makebox[0pt][r]{2021 IEEE International Conference on Big Data and Smart Computing}
}}}%

%
\title{Input Bias in Rectified Gradients and Modified Saliency Maps}


\author{\IEEEauthorblockN{Lennart Brocki}
\IEEEauthorblockA{Institute of Theoretical Physics, University of Wroclaw\\
Wroclaw, Poland\\
Institute of Informatics, University of Warsaw\\
Warsaw, Poland\\
Email: lennart.brocki@uwr.edu.pl }
\and
\IEEEauthorblockN{Neo Christopher Chung}
\IEEEauthorblockA{Institute of Informatics, University of Warsaw \\
Warsaw, Poland \\
University of California, Los Angeles School of Medicine\\
Los Angeles, CA, United States \\
Email: nchchung@gmail.com}
}


%


\maketitle

\begin{abstract}

Interpretation and improvement of deep neural networks relies on better understanding of their underlying mechanisms. In particular, gradients of classes or concepts with respect to the input features (e.g., pixels in images) are often used as importance scores or estimators, which are visualized in saliency maps. Thus, a family of saliency methods provide an intuitive way to identify input features with substantial influences on classifications or latent concepts. Several modifications to conventional saliency maps, such as Rectified Gradients \cite{Kim2019} and Layer-wise Relevance Propagation (LRP) \cite{bach2015pixel}, have been introduced to allegedly denoise and improve interpretability. While visually coherent in certain cases, Rectified Gradients and other modified saliency maps introduce a strong input bias (e.g., brightness in the RGB space) because of inappropriate uses of the input features. We demonstrate that dark areas of an input image are not highlighted by a saliency map using Rectified Gradients, even if it is relevant for the class or concept. Even in the scaled images, the input bias exists around an artificial point in color spectrum. Our modification, which simply eliminates multiplication with input features, removes this bias. This showcases how a visual criteria may not align with true explainability of deep learning models.
\end{abstract}


%
\IEEEpeerreviewmaketitle

\section{Introduction}

\begin{figure}[t]
	\centering
	\includegraphics[width=\columnwidth]{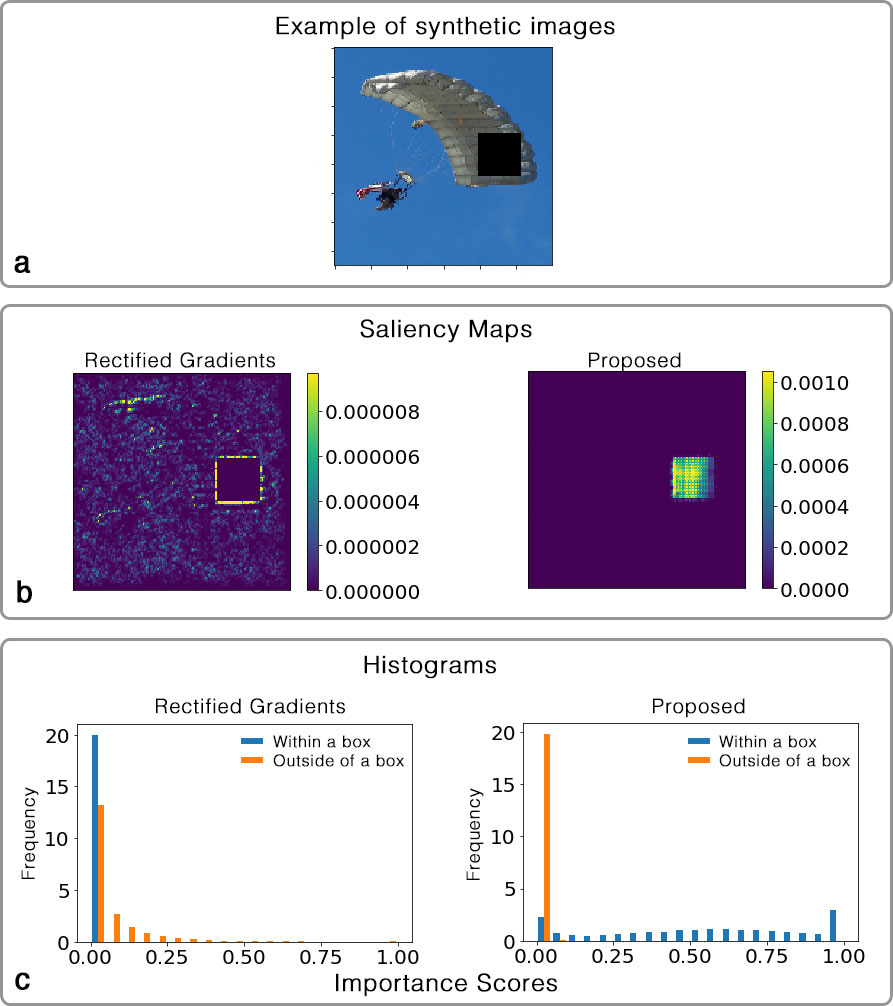}
	\caption{Simulation study demonstrating the strong brightness bias in Rectified Gradients. a) Synthetic examples were made from 4000 images from the Imagenette dataset. Black boxes were introduced to 2000 random images. b) Saliency maps were generated from Rectified Gradients and our proposed ``No Bias Rectified Gradients``.  The color-coding in the two saliency maps are different due to their intrinsic values. c) Importance scores are plotted in histograms comparing within and outside of a box. Note how for Rectified Gradient, all scores inside of the box are equal to zero.}
	\label{binary}
\end{figure}

While deep learning using artificial neural networks has shown great performance in computer vision and other areas, we are still seeking and developing ways to better explain why and how they work. Mechanisms underlying neural networks are often explored by saliency methods \cite{erhan2009visualizing, baehrens2010explain, Simonyan2014, Kim2019}. Generally, saliency methods calculate relevance or importance scores for input features (e.g., pixels in images) aimed at interpreting deep neural networks. Two-dimensional visualizations of importance scores are often referred to as saliency maps, which help to highlight input features relevant for classification or other tasks.

\begin{figure*}[!t]
	\centering
	\includegraphics[width=\textwidth]{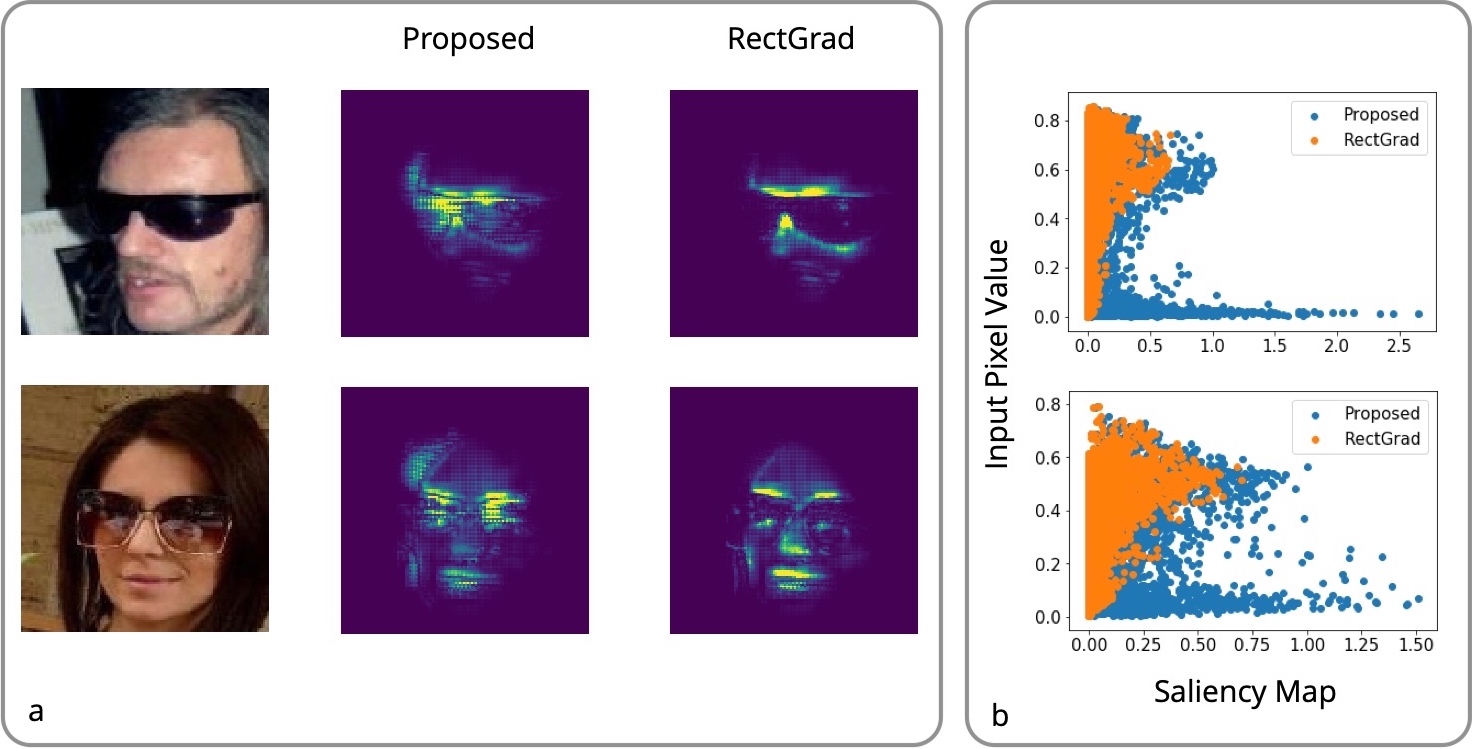}
	\caption{\small a) Saliency maps for ``eyeglasses'' using variational autoencoders (VAE). In saliency maps using ``RectGrad'' \cite{Kim2019}, the element-wise product of the input features and the gradient leads to being biased against the dark sunglasses that are not highlighted even though they are crucial. ``Proposed'' refers to our modification of Rectified Gradient, which removes the final multiplication step. b) Scatter plot of the pixel values (average over three color channels) in the original images and the saliency maps. RectGrad suppresses the gradients corresponding to dark pixels (e.g., 0), regardless of their importance for the sunglasses.}
	\label{bias_concept}
\end{figure*}

Saliency methods start from the gradient $M_{i}$ of a class score $S_\textnormal{class}$ with respect to the input features $p_{i}$, such as pixels for images $M_{i}=\partial S_\textnormal{class} / \partial p_{i}$. The class scores are usually taken to be activation of neurons in prediction vectors encoding the class of interest. The gradient tells us which input features have to be changed minimally to have a large influence on the class score. Therefore, larger magnitudes of gradients suggest greater relevance. 

Similarly, saliency methods can be applied to deep generative models such as Variational Autoencoder (VAE) \cite{Kingma2014}. To achieve applicability to VAE and other generative models, which naturally lack known classes, concept vectors are used to compute concept scores $S_\textnormal{concept}$, which can be understood as a replacement for $S_\textnormal{class}$  \cite{Brocki_2019}. A concept vector is a latent representation of a high-level concept, which could be known attributes \cite{Larsen2016, White2016}, cluster memberships \cite{Huang2016}, or others. The concept score $S_\textnormal{concept}$ is then obtained by measuring the similarity of the latent representation of an input image $\mathbf z_i$ and the concept vector corresponding to that attribute $\mathbf z_c$.

Calculation of gradients is central to saliency methods, and thus have been modified to produce visually sharper or de-noised visualization. Based on calculating gradients in successive layers of neural network, Guided Backpropagation thresholds any negative values of gradients in each layer during backpropagation \cite{Springenberg2015}. Rectified Gradients generalize this thresholding and is shown to result in sharper saliency maps in some cases \cite{Kim2019}. However, we notice an input bias in Rectified Gradients \cite{Kim2019}. Dark spots in an image were not highlighted by Rectified Gradients, irrespective of the relevance for a chosen class or concept. 

This bias is simply due to the element-wise product of the input features and the gradient. Similarly, variants of such multiplication steps, such as Layer-wise Relevance Propagation (LRP) \cite{bach2015pixel}, suffer from this bias. \cite{Adebayo2018} has shown how modified saliency maps showing unusual behaviors under randomization test. Our study provides a very simple reasoning behind why some modified saliency maps do not seem to deteriorate even when model parameters or data labels are randomized. For our concrete analysis, we focus on Rectified Gradient \cite{Kim2019} although our observation and modification are generally applicable. Comparisons are given using several neural networks, including both synthetic examples and real images. This straightforward study showcases how our visual inspection may lead underlying explainability of deep learning models astray.

A Python/Tensorflow package removing a bias step is available at \url{https://github.com/lenbrocki/nobias-rectified-gradient}.

\section{Bias in Rectified Gradient}\label{bias}

Gradients are generally noisy, when applied on image classification or other computer vision tasks involving deep neural networks \cite{erhan2009visualizing, baehrens2010explain, Simonyan2014, Kim2019}. Unlike how humans can easily identify an object in an image, it is not necessary for or constrained in neural networks to mimic this behavior. Rectified Gradient \cite{Kim2019} is a modified approach to the conventional calculation of gradients in order to reduce noise in saliency maps. Essentially, Rectified Gradients introduces layer-wise thresholding, such that artificially selected small values $< \tau$ are removed during backpropagation \cite{Kim2019}.

\begin{figure*}[t]
	\centering
	\includegraphics[width=\textwidth]{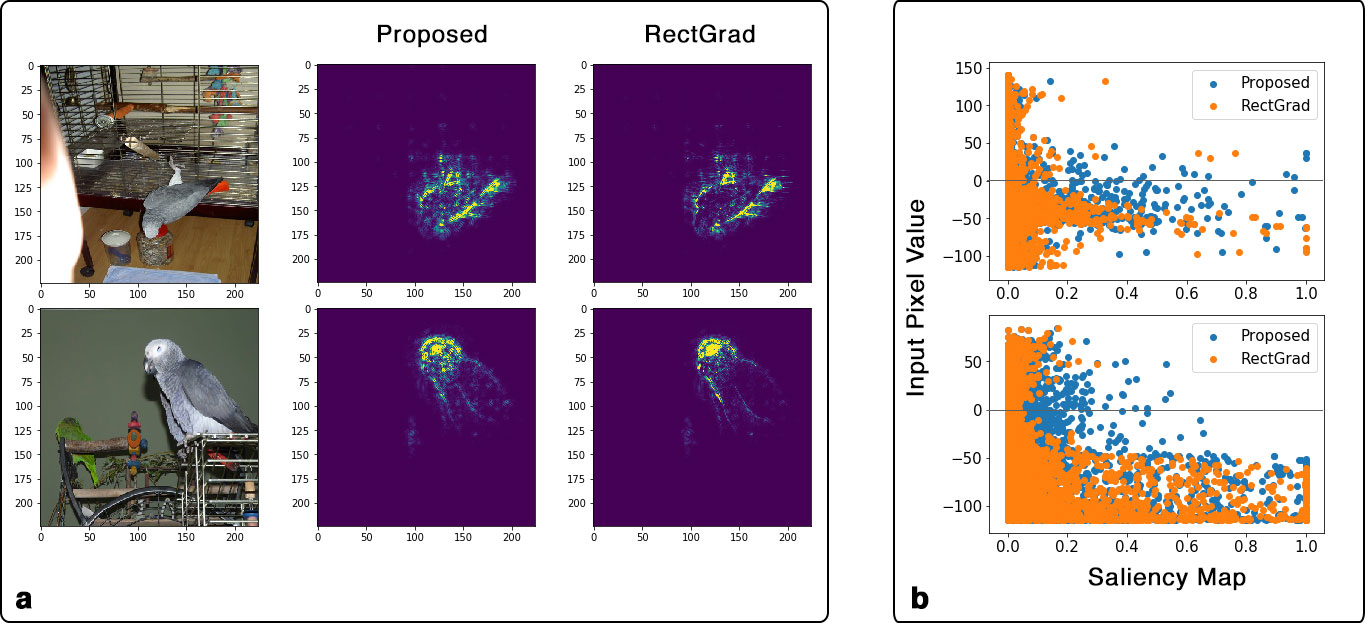}
	\caption{\small a) Saliency maps for the class ``grey great owls'' in ImageNet. In ImageNet, pixel values are centered around 0. Then, the element-wise product of the input features and the gradient in Rectified Gradient \cite{Kim2019} creates a bias against this artificial point in color spectrum. In this ImageNet example, a middle grey (and surrounding values) would be suppressed. We find this example with grey birds. b) Scatter plot of the pixel values (average over three color channels) in the original images and the saliency maps. Horizontal lines indicate the pixel value of 0 which are artificially suppressed in Rectified Gradients.}
	\label{bias_imagenet}
\end{figure*}

We have observed that Rectified Gradient introduces an input bias in the saliency maps. Due to this bias, dark areas of an image are not highlighted even if they are highly relevant for the classification. We have identified the source of this bias to be a final multiplication of the saliency map $M_i$ with the input features $x_i$ which is critical part of the definition of Rectified Gradient. The saliency methods are obtained by Rectified Gradient as follows
\begin{align}
	R_l^k &= \sum_{j}\mathbf{1}(a^k_l\cdot R^j_{l+1}>\tau)\cdot R^j_{l+1},\label{backprop}\\
	M_i &= x_i R_0^i,
\end{align}
where $\mathbf 1$ is the step function, $a^j_l$ is the activation $j$ in layer $l$, $R_l^j$ is the gradient backpropagated up to $a^j_l$ and $l=0$ is the input layer. The equation for the backpropagation of the gradients through the ReLU \eqref{backprop} is a modification of the rule obtained directly from the chain rule 
\begin{equation}
R_l^k = \sum_{j}\mathbf{1}(a^k_l > 0)\cdot R^j_{l+1},
\end{equation}
which is the definition used for the original saliency map \cite{erhan2009visualizing}. These modifications by \cite{Kim2019} were presented as to make the saliency maps less noisy. However, this has the undesirable effect of systematically neglecting dark features even when they are relevant. We therefore propose to simply remove the final multiplication with input features from Rectified Gradient, which will maintain its desirable feature of denoising the saliency maps while making it more generally applicable. This simplifies their method to $M_i = R_0^i$.

\section{Results}\label{result}

The consequence of multiplying the final gradients with input features is evident, since RGB (red, green, blue) values in images do not intrinsically contain meaningful information about classes or concepts. We demonstrate the resulting bias in three examples.

First, a simple synthetic example is generated by placing black squares on images from the Imagenette dataset (\url{https://github.com/fastai/imagenette}), a subset of ImageNet \cite{Russakovsky2015}. Out of 4000, 2000 images (50\%) were randomly chosen and black boxes were placed (Figure \ref{binary}(a)). As a result, the background images from the Imagenette dataset are not correlated to existence of black boxes. A binary classifier using three convolutional layers is built and trained on them. Training for 10 iterations achieved 99.8\% accuracy in the testing set, almost perfectly classifying the black box. Knowing that black boxes are artificially inserted into a random half of all images, only the black square in an image is relevant to this classifier. Thus, we expect the black square to be highlighted by having high values in accurate saliency maps. However, we noticed that Rectified Gradients of those black squares are zero, (Figure \ref{binary}(c)). Our proposed method successfully removes this bias.

Second, we investigate this bias when applying on variational autoencoders (VAE) on a large-scale face database with attributes called the CelebA dataset \cite{liu2015faceattributes}. All details about the architecture of the VAE and the training can be found in \cite{Brocki_2019}. Particularly, we focused on the concept ``eyeglasses'' which includes dark glasses and sunglasses. Figure \ref{bias_concept}(a) shows that our proposed method leads to a better highlighting (e.g., larger importance scores) of the dark glasses. Rectified Gradients (``RectGrad`` in Figure \ref{bias_concept}(a)) are suppressing any dark pixels and ended up with $M_i = 0$ if an input pixel is black $x_1 = 0$. Instead of highlighting the sunglasses, the saliency maps obtained using Rectified Gradient highlight the areas surrounding them. To show how gradients are modified, the scatterplot of the input features (averages of three color channels) and saliency maps (RectGrad or Proposed) are shown for each image (Figure \ref{bias_concept}(b)). Clearly, compared to our ``No Bias'' version, Rectified Gradients are showing artificially suppressed values when input pixel values are near 0. 

Third, we use a deep residual net with 50 layers of convolutional neural networks called ResNet50 \cite{he2016deep} to classify examples from ImageNet \cite{Russakovsky2015}. This example is used to demonstrate that preprocessing and normalization do not remove the observed bias. In some applications of deep neural networks, pixel values are scaled from $\lbrack 0, 255 \rbrack$ to $\lbrack -0.5, 0.5 \rbrack$ (or other ranges). This preprocessing only shifts this bias to an artificial point in the color spectrum. In particular, a range of $\lbrack -0.5, 0.5 \rbrack$ which is typically used in ImageNet classifiers would introduce this bias at a middle grey. Those pixels around R=127.5, G=127.5, B=127.5 (which correspond to R=0, G=0, B=0 in the scaled range) would be strongly suppressed and will not appear to have highly relevant values in Rectified Gradients. This bias is particularly observed in ImageNet classes whose objects are naturally grey (Figure \ref{bias_imagenet}(a)). We see that grey great owls are better represented by the proposed method. The scatterplot of the scaled input features and saliency maps demonstrate this artificial bias around 0 in the y-axis (Figure \ref{bias_imagenet}(b)).

\section{Discussion and Conclusion}

Interpretability of deep learning methods is highly sought after. While the saliency methods based on gradients are showing great potential to identify and visualize relevant input features, it is critical to carefully consider how our qualitative evaluation (e.g., visual inspection) may differ from the underlying mechanisms of classification or conceptualization.

We have compared the saliency maps obtained with Rectified Gradients and with our proposed ``No Bias'' modification, which removes the final multiplication with input pixels, for three different case studies. In the first synthetic example, the difference between the two methods is particularly apparent because the black box is completely removed from the saliency map by Rectified Gradient, whereas our proposed method correctly highlights it. The second example used the concept score method to obtain saliency maps for a VAE with respect to the concept ``eyeglasses''. The dark eyeglasses are substantially better highlighted when using our proposed method. In the third example, it is demonstrated that a preprocessing step of scaling or normalizing the pixel values does not remove this input bias. These artificial biases are consistently apparent when plotting the input pixel values against the importance scores in saliency maps.

Generally, we are interested in better understanding and explaining the predictions or underlying mechanisms of deep neural networks, that are independent of the human biases. One may argue that when the target object is bright, a multiplication with input pixels would lead to less noisy saliency maps. However, if we know the characteristics of the target object, such information should be incorporated into the model.

It can not be excluded that the saliency maps should be noisy, because this might accurately reflect the behavior of the model. In an extreme situation, the model may use a few pixels to reach its accurate prediction. Then, for many irrelevant pixels, the importance scores are highly noisy that will result in noisy saliency maps. If an artificially de-noised saliency map is desired, it is more transparent and interpretable to  post-processing of the maps.

\section*{Acknowledgment}

This work was supported by the Narodowe Centrum Nauki [2016/23/D/ST6/03613] and the NVIDIA Corporation’s GPU grant.



%

\bibliographystyle{IEEEtran}
\bibliography{refs}

%
%

\end{document}